\documentclass[letterpaper, 10 pt, conference]{ieeeconf}  

\IEEEoverridecommandlockouts                              
\usepackage{amsmath} 
\usepackage{graphicx}

\title{\LARGE \bf
Deep Classification of Epileptic Signals
}

\author{David Ahmedt-Aristizabal$^{1}$, Clinton Fookes$^{1}$, Kien Nguyen$^{1}$, Sridha Sridharan$^{1}$ 
\thanks{$^{1}$ David Ahmedt-Aristizabal, Prof. Clinton Fookes, Dr. Kien Nguyen and Prof. Sridha Sridharan are with Speech, Audio, Image and Video Technologies (SAIVT) research program, School of Electrical Engineering \& Computer Science, Queensland University of Technology, Brisbane, Australia.
}%
\thanks{Contact: david.aristizabal@hdr.qut.edu.au}%
}

\begin{document}

\maketitle
\thispagestyle{empty}
\pagestyle{empty}

\begin{abstract}

Electrophysiological observation plays a major role in epilepsy evaluation. However, human interpretation of brain signals is subjective and prone to misdiagnosis. Automating this process, especially seizure detection relying on scalp-based Electroencephalography (EEG) and intracranial EEG, has been the focus of research over recent decades. Nevertheless, its numerous challenges have inhibited a definitive solution. Inspired by recent advances in deep learning, we propose a new classification approach for EEG time series based on Recurrent Neural Networks (RNNs) via the use of Long-Short Term Memory (LSTM) networks. The proposed deep network effectively learns and models discriminative temporal patterns from EEG sequential data. Especially, the features are automatically discovered from the raw EEG data without any pre-processing step, eliminating humans from laborious feature design task. We also show that, in the epilepsy scenario, simple architectures can achieve competitive performance. Using simple architectures significantly benefits in the practical scenario considering their low computation complexity and reduced requirement for large training datasets. Using a public dataset, a multi-fold cross-validation scheme exhibited an average validation accuracy of 95.54\% and an average AUC of 0.9582 of the ROC curve among all sets defined in the experiment. This work reinforces the benefits of deep learning to be further attended in clinical applications and neuroscientific research.

\end{abstract}

\section{INTRODUCTION}

Epilepsy is a neurological disorder characterised by frequent and unpredictable seizures. Prior to epilepsy diagnosis, patients are usually monitored using a broad range of information from neuroimaging and electrophysiological methods~\cite{rosenow2001presurgical}. Electroencephalography (EEG) has long been considered a gold standard for the diagnosis of seizures. The goal of the epilepsy evaluation is to delineate the brain network affected. However, this network could comprise other networks, which are involved in originating interictal epileptiform discharges and producing the first clinical manifestation of a seizure~\cite{chauvel2014emergence}. Misjudgement of the location of these networks causes ineffective clinical decisions. 

Despite recent advances in developing automated seizure detection devices~\cite{ahmedt2017automated}, none of them is universally accepted because the performance in clinical scenarios has not been satisfactory. Significant work is still needed to reach expert-level evaluation, especially in understanding the epileptiform activities~\cite{wendling2016computational,antoniades2016deep}, and by generalizing representations that are invariant to inter- and intra-subject differences. The performance of the traditional detection approaches relies heavily on expert knowledge to design the signal features employed and regularly include frequency-based features such as the wavelet transform, and energy analysis~\cite{ramgopal2014seizure}. However, there is no warranty that these hand-crafted features are optimal for the chosen task, especially in the complex scenario of brain electrical activity. A major question to be asked is whether the feature engineering can be conducted automatically to discover features directly from the data, and not avoid the use of human expect knowledge, domain knowledge, and human bias. 

Recent advances in deep learning could be the answer to this question. Deep learning is a subset of the machine learning family which simulates structures and operations of a human brain through a hierarchical multiple-layer signal representation coupled with advanced training algorithms~\cite{lecun2015deep}. The major advances of deep learning in comparison with traditional machine learning is that the spatial, spectral and temporal feature representation is automatically learned from the training data, not by human assumption, leading to natural and effective signal representation and superior performance~\cite{lecun2015deep}. 
Deep learning has revolutionised many computer vision and medical applications, especially for classifying brain signals~\cite{ahmedt2017automated,bozhkovoverview,schirrmeister2017deep}.
Additionally, these deep architectures have transformed the tasks of seizure detection during the processing of EEG recordings for epilepsy diagnosis
~\cite{thodoroff2016learning,vidyaratne2016deep,page2016wearable,lin2016classification}.
Despite their benefits, there are two major limitations of these existing approaches: they either pre-process the raw data into some other forms before being fed into a CNN; or they involve the use of very deep and complex networks which have millions of parameters to be trained~\cite{acharya2017deep}. and require very large training datasets, which are usually not available in the clinical scenarios. 

In order to address these limitations, we investigate the plausibility of simple deep learning architectures that are capable of both abstracting high-order features and classifying them according to the physiological brain state and achieve state-of-the-art performance. A simple network will represent the benefits of deep models to achieve high performance with fast run-time and reduced need for large datasets, as it is noteworthy that seizure events are rare as it is in clinical scenarios. In addition, the proposed networks process the raw data directly, without any transformation to the original EEG recordings and exploit the temporal patterns through the use of LSTMs. By automatically exploiting and discovering features from the temporal information, the proposed networks could extract robust and reliable patterns to classify epileptic signals. 

\begin{figure*}
\begin{center}
\includegraphics[width=0.9\linewidth]{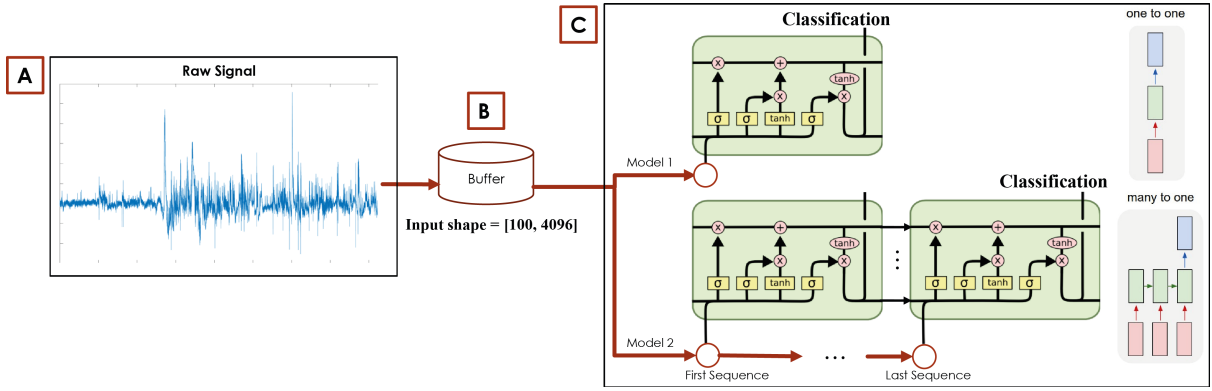}
\end{center}
   \caption{Block diagram of the proposed deep framework. A. The raw samples for each type of brain-state are concatenated without any pre-processing. The value of amplitude of the signal is considered as a single representation of the segment size. B. The temporal evolution of the signal is analysed using the complete length of the signal, which indicates a total number of 4,096 segments and 100 samples for each type. C. The feature sequence is feed to differents models based on Long-Short-Term-Memory (LSTM) architectures to exploit the temporal relation between segments and to predict brain-states signals.}
\label{fig:Fig1}
\end{figure*}

The remainder of this paper is organised as follows: Section II presents the dataset and the methodology, Section III illustrates and discusses the results. Finally, Section IV reviews and concludes the paper.

\section{MATERIALS AND METHODS}

\subsection{Dataset}
The experimental data is from the publicly available dataset from the Department of Epileptology, University of Bonn~\cite{andrzejak2001indications}. The dataset includes five sets (denoted from A to E) with a total of 100 EEG samples for each set. Each sample is a single channel EEG recorded at 173.6 Hz with 23.6 seconds of duration. Thus, the sample length of each sample is 4,096. Set A and B were recorded using scalp EEG from five healthy volunteers (healthy state) with eyes open and closed respectively. Set C, D and E, from five epileptic patients prior to surgery diagnosed with Temporal Lobe Epilepsy, were recorded using depth electrodes implanted symmetrically into the hippocampal formation. Set C and D were during seizure-free intervals, where set D was recorded from the epileptogenic zone (Inter-Ictal state or between seizures) and set C from the opposite brain hemisphere. Finally, set E described the recordings of the epileptogenic zone during an epileptic seizure (Ictal state).

\subsection{Method}
The aim of this investigation is to compare properties of brain electrical activity from different recording regions and from different pathological brain states, i.e. classify healthy, inter-ictal and ictal EEG signals. To achieve this, we propose a deep framework which receives the raw EEG signals and extracts temporal features using an end-to-end deep learning model known as Long Short-Term Memory (LSTM) architecture~\cite{hochreiter1997long} illustrated in Fig. \ref{fig:Fig1}. An LSTM is a special type of Recurrent Neural Network (RNN), especially suited for sequential data such as EEG signals since their neurons contain connections (weights) not only between the successive layers but also to themselves, which are used to memorise information from previous inputs. Therefore, LSTM networks are capable of learning long-term dependencies present in sequential data and could predict time series when there are long time lags of unknown size between important events~\cite{greff2017lstm}.

RNNs are called recurrent because they perform the same task for every element of a sequence, with the output being depended on the previous computations. Another way to think about RNNs is that they have a memory which captures information about what has been calculated so far. However, RNNs perform poorly when dealing with long sequences due to its famous drawback in gradient vanishing and exploding. LSTMs seek to address this issue by using a gated mechanism. Three gates, i.e. forget, input and output gates, are used to control the flow of information. The amount of information that is let through each gate is controlled by a point-wise multiplication and sigmoid function. For a system with input $x_t$, an output $y_t$ and a hidden state $h_t$, a conventional RNN is constructed by defining the transition function and the output function as,

\begin{equation}
h_{t} = \phi_{b} (W^{T}h_{t-1}+U^{T}x_{t}) ; 
y_{t} = \phi_{o} (V^{T}h_{t})
\end{equation}

where W, U and V are the transition, input and output matrices respectively and $\phi_{b}$ and $\phi_{o}$ are element-wise nonlinear functions. Sigmoid or a hyperbolic tangent function are common examples of nonlinear functions. When the forget and input gates have determined how much information of the previous cell state $C_{t-1}$ and the new cell state candidate $\hat{C}_t$  should be let through, the dynamic equations to represent the LSTM is given as,

\begin{equation}
\begin{split}
\hat{C}_{t} = tanh(W^{T}(r_t*h_{t-1})+U^{T}x_{t}) \\
z_{t} = \sigma_{b}(W^{T}_{z}h_{t-1}+U^{T}_{z}x_{t}+V^{T}_{z}C_{t-1}) \\
C_{t} = f_{t}*C_{t-1}+i_{t}*\hat{C}_{t} \\
h_{t} = o_{t}*\phi_{b}(C_{t})
\end{split}
\end{equation}

where $z=\big\{i,f,o,r\big\}$, representing the gating functions: input gate, the forget gate, the output gate and the internal gate, and σ is the Sigmoid function. The trainable model parameters are: $\big\{W,W_z,U,U_z,V_z\big\}$. 

\begin{table}[h]
\caption{LSTM ARCHITECTURES}
\label{table1}
\begin{center}
\begin{tabular}{|c|c|c|c|c|}
\hline
 & \multicolumn{2}{c}{\bf \ Model 1: One to One} & \multicolumn{2}{|c|}{\bf \ Model 2: Many to One} \\ \hline
Layer Type & Output & Parameters & Output & Parameters \\ \hline
Input & (4097,1) &  & (4097,1) &  \\
$LSTM_1$ & (4097,1) & 16,896 & (4097,128) & 66,560 \\
$Dropout_1$ &  &  & (4097,128) & \\
$LSTM_2$ &  &  & (64) & 49,408 \\
$Dropout_2$ &  &  & (64) &  \\
$Dense_1$ & (-1) & 65 & (1) & 65 \\ \hline
\textbf{Total} & & \textbf{16,961} & & \textbf{116,033} \\ \hline
\end{tabular}
\end{center}
\end{table}

\subsection{Architecture and Classification} 
The block diagram of the proposed deep learning system is presented in Fig. \ref{fig:Fig1}. Each EEG sequence comprises of 4096 segments. The complete temporal sequence for each set has an input shape of $[100,4096]$. This illustrates 100 samples, each of them with 4096 segments. The representation of each set is saved in a buffer; then used as the input to an LSTM to exploit temporal information of each EEG signal. Unlike the previous proposals where the features are hand-crafted and the signals are pre-processed, our method automatically learns the inherent characteristics of seizure data. The classification verifies whether a validation or test signal has similar dynamical properties of the brain-states. 
The number of LSTM layers is one significant hyper-parameter to consider in the LSTM network. We have experimented with various implementations with different numbers of layers, ranging from 1 single layer to multiple layers. The model one-to-one indicates that from one single layer, the model estimates one single output. On the other hand, the model many-to-one, refers to multiple stack LSTMs that infer one output. The architectures are selected according to the performance of the model for each pair-set. Table \ref{table1} describes the specific LSTM models adopted in the classification process. We experimented with various numbers of memory cells in each layer and obtained the best performance with a network configured with one single layer with 64 hidden units (Model 1) and with 2 hidden layers of 128 and 64 hidden units respectively (Model 2). The output of the recurrent layer is fed into a densely-connected neural network layer with a sigmoid activation function to predict the set probability for the input data sequence. We also tested more complex architectures but the performance gain is not significant. More complicated architectures have more capability to model complicated signals, but practical clinical implementation would be affected; hence simple architectures with one or two layers could yield very accurate results in the
experimental data. Therefore, the models are lightweight, with on the order of less than 17,000 trainable parameters in the case of Model 1.

Training is carried out by optimizing the binary cross entropy loss function. The LSTM is optimised with the ADAM optimizer with a learning factor of $10^{-3}$, and decay rate of first and second moments as 0.9 and 0.999 respectively. Batch size set to 4 and dropout with a probability of 0.35, for Model 2, are considered to reduce the overfitting because of the small amount of training data available in the dataset. We perform the model training using only 20 epochs and use the default initialization parameters from the Keras package for initializing the weights of the LSTM hidden units.

\section{EXPERIMENTAL RESULTS}

The proposed networks were employed to classify six pairs of EEG recordings. These pairs are illustrated in Table \ref{table2}. For instance, the classification between set A and E refers to the verification between healthy and seizure EEG signals. 

In the experiment, we adopted a k-fold cross-validation to verify the generalization and robustness of the proposed architecture, where the LSTM architectures were fixed for each validation. The k-fold cross-validation~\cite{kohavi1995study} is one way to confirm the reliability of the model to predict data that has not already been seen. In this method, the samples of each set are randomly split into 70\% for training, 20\% for validation and 10\% for testing k different folds (10-folds in this experiment). The difference between the validation and test samples is that the last one is not seen during the training phase. The performance of our method in each pair-set is assessed by the average of the best result of each fold with the metrics of validation and test accuracy, sensitivity, specificity, precision and the area under the curve (AUC). The multi-fold cross-validation average performance is displayed in Table \ref{table2}. The deep framework was capable of achieving an average of 95.54\% in the validation accuracy and an average area under the curve of 0.9582 between all the sets pairs. The validation accuracy and error over the training process are shown in Fig. 3 and Fig. 4, respectively. This demonstrates that the learned features showed clear differences in dynamical properties of brain electrical activity from different physiological brain states.

\begin{table*}[h]
\caption{MULTI-FOLD CROSS-VALIDATION PERFORMANCE (10-TIME AVERAGE)}
\label{table2}
\begin{center}
\begin{tabular}{|c|p{1cm}|p{1.5cm}|p{1.5cm}|p{1.5cm}|p{1.5cm}|p{1.5cm}|c||p{1.5cm}|}
\hline
Sets & Type of Model & Validation Accuracy (\%) & Test Accuracy (\%)
 & Test Sensitivity (\%) & Test Specificity (\%) & Test Precision (\%) & AUC & Validation Accuracy (\%)~\cite{lin2016classification} 
 \\ \hline
A and E & 1 & \textbf{99.50} & \textbf{97.00} & 96.00 & 98.00 & 98.09 & \textbf{0.9820} & \textbf{95.50}\\
B and E & 1 & 94.75 & 92.50 & 91.00 & 94.00 & 94.27 & 0.9850 & 92.50\\
C and E & 1 & 97.25 & 92.00 & 95.00 & 89.00 & 90.06 & 0.9650 & 91.67\\
D and E & 1 & 96.50 & 91.00 & 95.00 & 87.00 & 89.06 & 0.9510 & 93.34\\
A and D & 2 & \textbf{90.25} & \textbf{82.00} & 82.00 & 82.00 & 84.78 & \textbf{0.9030} & \textbf{86.42}\\
B and D & 2 & 95.00 & 93.00 & 92.00 & 93.00 & 93.00 & 0.9630 & N.A \\ \hline
\textbf{Average} & & \textbf{95.54} & \textbf{91.25} & 91.83 & 90.50 & 91.50 & 0.9582 &  \\ \hline
\end{tabular}
\end{center}
\end{table*}

We can see that the proposed framework achieves a significantly high accuracy of classification with simple deep learning architectures, which means low computational cost. As illustrated the Table II, this performance outperformed the results reported in the literature and those documented in [13], where the 90\% of the data was used for training. In our method, the highest accuracy is obtained with the pair of Set A and E, while the lowest with Set A and D. This result was expected because the dynamical properties of the signals from the epileptogenic zone between seizures are more similar to healthy EEG segments than to ictal signals.

\section{CONCLUSIONS}

We have investigated the benefits of a recurrent deep learning framework to classify EEG segments from epileptic signals. We adopt LSTM networks to extract temporal patterns in the frame sequences. From the experimental results, simple models can achieve a very high degree of accuracy. The proposed approach demonstrates the capability of recurrent models to learn a general representation of a seizure event from the raw data automatically, which could enhance the diagnosis and treatment planning for patients that experience epilepsy.

\begin{figure}[t]
\begin{center}
\includegraphics[width=0.9\linewidth]{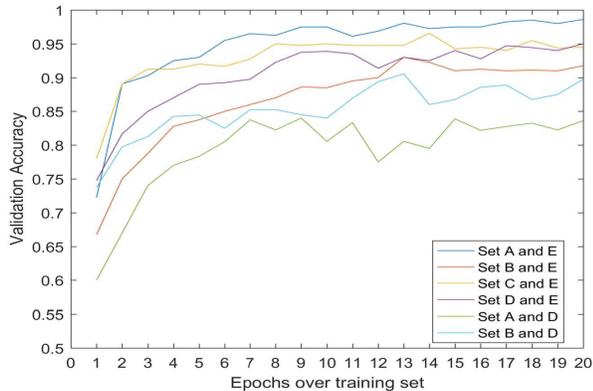}
\end{center}
   \caption{Validation accuracy performance of all sets. (Best in color).}
\label{fig:long}
\label{fig:onecol}
\end{figure}

\begin{figure}[t]
\begin{center}
\includegraphics[width=0.9\linewidth]{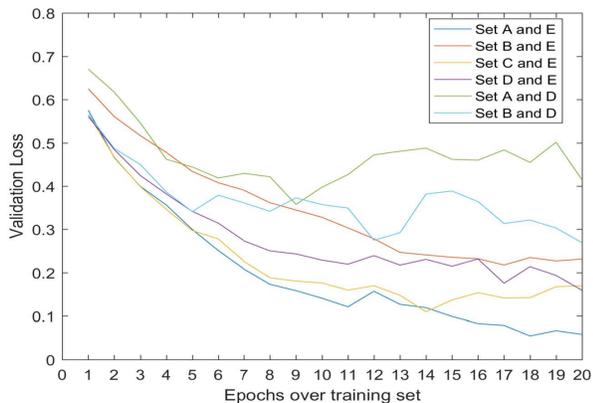}
\end{center}
   \caption{Validation error performance of all sets. (Best in color).}
\label{fig:long}
\label{fig:onecol}
\end{figure}

\section{ACKNOWLEDGEMENT}
The authors acknowledge the support of the QUT High Performance Computing (HPC) for providing the computational resources for this research.

\addtolength{\textheight}{-12cm}   





\bibliographystyle{IEEEtran}
\bibliography{egbib}

\end{document}